\documentclass[conference]{IEEEtran}
\IEEEoverridecommandlockouts
\usepackage{cite}
\usepackage{amsmath,amssymb,amsfonts}
\usepackage{algorithmic}
\usepackage{graphicx}
\usepackage{textcomp}
\usepackage{xcolor}
\usepackage{tikz}
\def\BibTeX{{\rm B\kern-.05em{\sc i\kern-.025em b}\kern-.08em
    T\kern-.1667em\lower.7ex\hbox{E}\kern-.125emX}}
    
\usepackage[pagebackref=true,breaklinks=true,colorlinks,bookmarks=false]{hyperref}
\usepackage[nameinlink]{cleveref}

\newcommand{\IoU}{\mathit{IoU}}

\DeclareMathOperator*{\argmax}{arg\,max}
\DeclareMathOperator*{\argmin}{arg\,min}

\usepackage{subfigure}
\definecolor{slateblue}{rgb}{0.42, 0.35, 0.8}
\definecolor{indianred}{rgb}{0.8, 0.36, 0.36}
\definecolor{mediumseagreen}{rgb}{0.24, 0.7, 0.44}
\definecolor{darkgoldenrod}{rgb}{0.72, 0.53, 0.04}
\definecolor{mediumred-violet}{rgb}{0.73, 0.2, 0.52}
\definecolor{cornflowerblue}{rgb}{0.39, 0.58, 0.93}
\definecolor{goldenyellow}{rgb}{1.0, 0.87, 0.0}
\definecolor{darkseagreen}{rgb}{0.56, 0.74, 0.56}
\definecolor{lightskyblue}{rgb}{0.53, 0.81, 0.98}
\definecolor{mediumslateblue}{rgb}{0.48, 0.41, 0.93}
\definecolor{plum(web)}{rgb}{0.8, 0.6, 0.8}
\definecolor{mediumorchid}{rgb}{0.73, 0.33, 0.83}

\begin{document}

\title{Detection of False Positive and False Negative Samples in Semantic Segmentation
\thanks{Work supported by Volkswagen Group Innovation and BMWi grant no. 19A119005R.}
}

\author{
\IEEEauthorblockN{Matthias Rottmann}
\IEEEauthorblockA{
University of Wuppertal \& ICMD\\
rottmann@math.uni-wuppertal.de}\\  
\IEEEauthorblockN{Fabian H\"uger}
\IEEEauthorblockA{
Volkswagen Group Innovation\\
fabian.hueger@volkswagen.de}
\and
\IEEEauthorblockN{Kira Maag}
\IEEEauthorblockA{University of Wuppertal \& ICMD\\
maag@math.uni-wuppertal.de}\\
\IEEEauthorblockN{ Peter Schlicht}
\IEEEauthorblockA{Volkswagen Group Innovation\\
peter.schlicht@volkswagen.de}
\and
\IEEEauthorblockN{Robin Chan}
\IEEEauthorblockA{University of Wuppertal \& ICMD\\
chan@math.uni-wuppertal.de}\\              
\IEEEauthorblockN{ Hanno Gottschalk}
\IEEEauthorblockA{University of Wuppertal \& ICMD\\
hanno.gottschalk@uni-wuppertal.de}
}

\newcommand{\commentMR}[1]{#1}
\newcommand{\commentKM}[1]{#1}
\newcommand{\commentRC}[1]{#1}
\newcommand{\commentFH}[1]{#1}
\newcommand{\commentPS}[1]{#1}
\newcommand{\commentHG}[1]{#1}
\newcommand{\changed}[1]{#1}
\newcommand{\comment}[1]{#1}

\maketitle

\begin{abstract}
In recent years, deep learning methods have outperformed other methods in image recognition. This has fostered imagination of potential application of deep learning technology including safety relevant applications like the interpretation of medical images or autonomous driving. The passage from assistance of a human decision maker to ever more automated systems however increases the need to properly handle the failure modes of deep learning modules. In this contribution, we review a set of techniques for the self-monitoring of machine-learning algorithms based on uncertainty quantification. In particular, we apply this to the task of semantic segmentation, where the machine learning algorithm decomposes an image according to semantic categories. We discuss false positive and false negative error modes at instance-level and review techniques for the detection of such errors that have been recently proposed by the authors. We also give an outlook on future research directions.    
\end{abstract}


\section{Introduction}
The stunning success of deep learning technology, convolutional neural networks (CNN) in particular \cite{lecun1995comparison,lecun2015lenet,goodfellow2016deep}, has led to a rush towards technology development for new applications that ten years ago would have been considered unrealistic. In particular, fully automated driving systems are intensively developed in the automotive industry including also new competitors \cite{berger2014autonomous,smith2015automated}. While the industry strives to advance such systems from driving assistance for a human driver (level 1 and 2 of automated driving) to higher levels where the human as the ultimate redundancy for the technology can be temporally (level 3 and 4) or entirely replaced (level 5), the question of how to design automated driving systems based on deep learning technology still poses a number of unresolved questions, in particular with respect to reliability and safety \cite{smith2015automated,safetyfirst2019}. A similar set of problems exists when AI-driven systems assist the interpretation of medical images \cite{kermi2018deep}, although there is no intention to fully automate this process.    

In the following, we focus on the semantic interpretation of street scenes based on camera data which is an important prerequisite for any automated driving strategy. For the sake of concreteness, we focus on semantic segmentation in contrast to object detection \cite{redmon2016you}. In semantic segmentation, an image is decomposed into a number of masks, each of which unifies the pixels that adhere to a specific category in a predefined semantic space\cite{Guo2018}. Despite there also exist instance segmentation networks \cite{girshick2014rich}, we here consider each connected component of a mask as one instance. Based on such instances, the following failure modes have to be taken into account:
\begin{itemize}
\item False Positive (FP): An instance of a given category that is present in the predicted mask has zero intersection with the same category in the ground truth mask. 
\item False Negative (FN): An instance of a given category that is present in the ground truth mask is completely overlooked, i.e., has zero intersection with the same category in the predicted mask. 
\item Out of Distribution (OOD): An object that is outside the semantic space on which the perception algorithm has been trained nevertheless occurs in the input data and therefore is misclassified \cite{liang2017enhancing,meinke2019towards}.
\item Adversarial Attack (AA): The perception module is intentionally forced to commit an FP or FN error by manipulation of the input of the sensor \cite{Szegedy2013IntriguingPO,goodfellow2016deep}.   
\end{itemize}           
In the following, we focus on the first two 'FP' and 'FN' failure modes. In particular, we discuss methods for 
self-monitoring of segmentation networks. 
Improved reliability due to redundancies in the architecture of autonomous cars is not considered here.   

While the detection of false positives in semantic segmentation is mostly considered on a pixel level and is measured with global indices like the global accuracy over frames or the averaged intersection over union (IoU) on class mask level \cite{xception,mobilenet}, here we pass on to connected components in the predicted masks of segmentation networks, which is often more relevant in practice. Meta classification then is the machine learning task to infer from the aggregated uncertainty metrics whether the predicted segment has intersection with the ground truth, or is a false positive in the sense given above. While this results in a 0 -- 1 decision, the IoU score on \emph{a single connected component} gives a gradual quality measure. Meta regression then is the task to predict this score from the same uncertainty metrics in the absence of 
ground truth. In this article we give an overview over recent progress in meta classification \cite{Rottmann2018,Maag2019,Schubert2019} for the semantic segmentation of street scenes \cite{Geiger2013}.

We also deal with class imbalance as one of the reasons for false negative predictions for which the corresponding ground truth is 
underrepresented in the training data. In semantic segmentation this is often unavoidable, as e.g.\ pedestrians are underrepresented in terms of their pixel count even on images with several individuals. Here we propose methods to correct the bias from the maximum \emph{a posteriori} (or Bayes) decision principle that is mostly applied in machine learning. As alternatives we propose a decision principle -- the maximum likelihood (ML) decision rule \cite{Fahrmeir1996} -- that looks out for the best fit of the data to a given semantic class. We review the false negative detection \cite{Chan2019} using the ML decision rule for semantic segmentation and also discuss cost based decision rules in general along with the problems of setting the cost structure up \cite{Chan2019Dilemma}.

The paper is organised as follows: In section \ref{sec:FP} we discuss the detection of false positive instances by a meta classification procedure that involves uncertainty heatmaps aggregated over predicted segments. The following section \ref{sec:TDY} extends this procedure to video stream data using time series of uncertainty metrics for meta-classification. In section \ref{sec:DR} we discuss the reduction of false negatives especially for rare classes of high importance. Here we use cost based decision rules and discuss some of the ethical issues connected with setting up the cost structure. Finally, section \ref{sec:OL} gives a summary and outlook to future research.

\section{\label{sec:FP}False Positive Detection via Meta Classification}


In semantic segmentation, a network with a softmax output layer provides for each pixel $z$ of the image a probability distribution $f_z(y|x,w)$ on the $q$ class labels $y\in\mathcal{C}=\{y_1,\ldots,y_q\}$, given the weights $w$ and input image $x \in \mathcal{X}$. Using the maximum \emph{a posteriory} probability (MAP) principle, also called Bayes decision rule, predicted class in $z$ is then given by
\begin{equation}\label{eq:map}
\hat y_z(x,w)=\argmax_{y\in\mathcal{C}}f_z(y|x,w).
\end{equation}
We denote by $\hat{\mathcal K}_x$ the set of connected components (segments) in the predicted segmentation $\hat{\mathcal S}_x=\{\hat y_z(x,w) | z\in x\}$. Analogously we denote by ${\mathcal K}_x$ the set of connected components in the ground truth ${\mathcal S}_x$. 

Let ${\mathcal K}_x|_k$ be the set of all $k'\in {\mathcal K}_x$ that have non-trivial intersection with $k$ and whose class label equals the predicted class for $k$, then the intersection over union ($\IoU$) is defined as
\begin{equation}
    \IoU(k) = \frac{|k \cap K'|}{|k \cup K'|}\,,\qquad K' = \bigcup_{k' \in {\mathcal K}_x|_k} k'.
\end{equation}
In the given context, false positive detection (cf.~\cite{HendrycksG16c} for classification tasks with neural networks) corresponds to the binary classification task $\IoU(k)=0$ or $\IoU(k)>0$ for a given segment $k \in \hat{\mathcal K}_x$, i.e., $k$ intersect with ground truth or not. We term this task \emph{meta classification}, see \cite{Rottmann2018}.
In analogy we term the regression task of estimating $\IoU(k)$ directly as \emph{meta regression}, this can also be viewed as a quality measure. Quality estimates for neural networks were first proposed for one object per image in~\cite{seg_quality,qualitynet}. While these works rely on neural networks as post processors, we introduced a light-weight and transparent approach in \cite{Rottmann2018} that deals with multiple segments per image for both meta tasks.

\begin{figure*}[t]
    \centering
    \includegraphics[width=.66\linewidth]{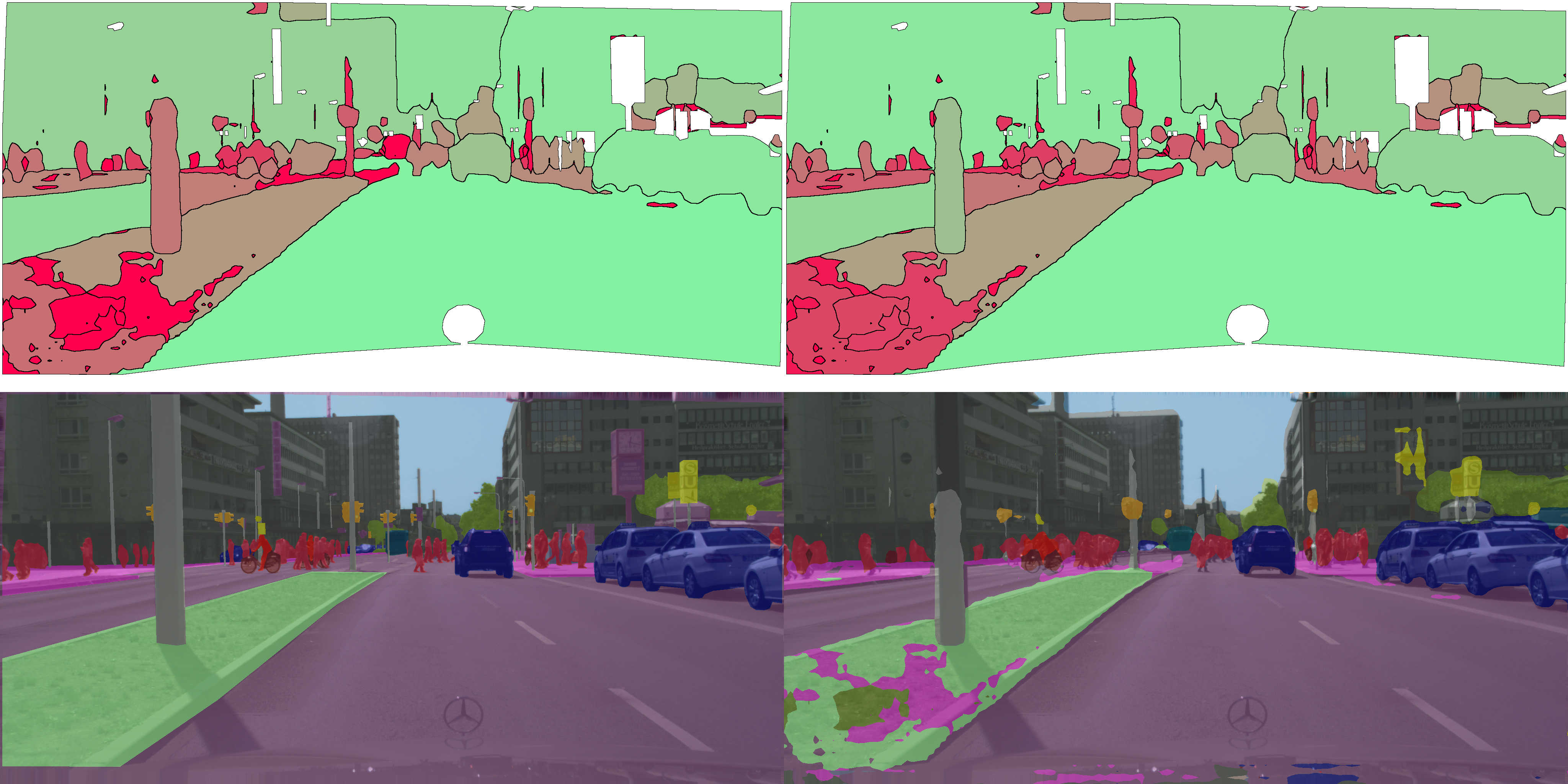}
    \caption{Prediction of the $\IoU$ with linear regression. The figure consists of ground truth (bottom left), predicted segments (bottom right), true $\IoU$ for the predicted segments (top left) and predicted $\IoU$ for the predicted segments (top right). In the top row, green color corresponds to high $\IoU$ values and red color to low ones, for the white regions there is no ground truth available. These regions are excluded from the statistical evaluation.}
    \label{fig:metaseg}
\end{figure*}

In our approach presented in \cite{Rottmann2018} we proceed as follows: for each $k \in \hat{\mathcal K}_x$ we construct metrics based on dispersion measures of $f_z(y|x,w)$ (entropy, probability margin) as well as fractality measures of $k$. The dispersion measures are aggregated over the predicted segment by computing their averages, fractality is measured by the quotient of volume and boundary length of $k$. We observe that these metrics are strongly correlated with the $\IoU(k)$, yielding Pearson correlation coefficients $R$ of up to 0.85 (in absolute values) for two different state-of-the-art DeepLabv3+ \cite{deeplab} networks (Xception65 \cite{xception} and MobilenetV2 \cite{mobilenet}). Hence, the constructed metrics are suitable for both meta tasks. The construction of metrics can be seen as a map $\mu: \hat{\mathcal K}_x \to \mathbb{R}^m$ that maps $k$ to a vector of metrics. Thus, \begin{equation}
    M = \{ \mu(k) \, : \, x \in \mathcal{X}, \; k \in \hat{\mathcal K}_x \}, \; M_i = \{ \mu_i(k) \}
\end{equation} is a structured dataset, for further details on the construction of $M$ we refer to \cite{Rottmann2018}.
We perform meta tasks by training linear models, i.e., a linear regression model for meta regression and a logistic one for meta classification, both of them based on $M$. We split the set of all predicted segments and their corresponding metrics obtained from the Cityscapes \cite{cityscapes} validation set into meta training and meta test sets (80\%/20\%) and compare our approach with the following baselines: for the entropy baseline we employ for both meta tasks a single metric, i.e., the mean entropy over a predicted segment $k$ as the entropy is a commonly used uncertainty measure. Furthermore, for the classification task a naive random guessing baseline can be formulated by randomly assigning a probability to each segment $k$ and then thresholding on it. A comparison of our meta classification approach is given in \cref{tab:summary1}. Noteworthily, we obtain AUROC values of up to $87.72\%$ (roughly 10 percent points (pp.) above the entropy baseline) for meta classification and $R^2$ values of up $81.48\%$ (more than 30 pp.\ above the entropy baseline) for meta regression. A visualization demonstrating the performance of our approach is given in \cref{fig:metaseg}. In \cite{Rottmann2018}, we also present results for the BraTS2017 brain tumor segmentation dataset \cite{brats2017}.

\begin{table}[t]
\begin{center}
\caption{Summarized results for classification and regression for Cityscapes, averaged over 10 runs. The numbers in brackets denote standard deviations of the computed mean values.} \label{tab:summary1}
\scalebox{0.65}{
\begin{tabular}{|l||r|r||r|r|}
\hline
& \multicolumn{2}{|c||}{Xception65} & \multicolumn{2}{|c|}{MobilenetV2}  \\
\hline
Cityscapes & \multicolumn{1}{|c|}{training} & \multicolumn{1}{|c||}{validation} & \multicolumn{1}{|c|}{training} & \multicolumn{1}{|c|}{validation} \\
\hline
& \multicolumn{4}{|c|}{Meta Classification $\IoU=0,>0$} \\
\hline
ACC, penalized             & $81.88\%(\pm0.13\%)$ & $81.91\%(\pm0.13\%)$ & $78.87\%(\pm0.13\%)$ & $78.93\%(\pm0.17\%)$ \\
ACC, unpenalized           & $81.91\%(\pm0.12\%)$ & $81.92\%(\pm0.12\%)$ & $78.84\%(\pm0.14\%)$ & $78.93\%(\pm0.18\%)$ \\
ACC, entropy only      & $76.36\%(\pm0.17\%)$ & $76.32\%(\pm0.17\%)$ & $68.33\%(\pm0.27\%)$ & $68.57\%(\pm0.25\%)$ \\
\hline
ACC, naive baseline        & \multicolumn{2}{|c||}{$74.93\%$} & \multicolumn{2}{|c|}{$58.19\%$} \\
\hline
AUROC, penalized           & $87.71\%(\pm0.14\%)$ & $87.71\%(\pm0.15\%)$ & $86.74\%(\pm0.18\%)$ & $86.77\%(\pm0.17\%)$ \\
AUROC, unpenalized         & $87.72\%(\pm0.14\%)$ & $87.72\%(\pm0.15\%)$ & $86.74\%(\pm0.18\%)$ & $86.76\%(\pm0.18\%)$ \\
AUROC, entropy only    & $77.81\%(\pm0.16\%)$ & $77.94\%(\pm0.15\%)$ & $76.63\%(\pm0.24\%)$ & $76.74\%(\pm0.24\%)$ \\
\hline
& \multicolumn{4}{|c|}{Meta Regression $\IoU$} \\
\hline
$\sigma$, all metrics      & $0.181(\pm0.001)$    & $0.182(\pm0.001)$    & $0.130(\pm0.001)$    & $0.130(\pm0.001)$    \\ 
$\sigma$, entropy only     & $0.258(\pm0.001)$    & $0.259(\pm0.001)$    & $0.215(\pm0.001)$    & $0.215(\pm0.001)$    \\ 
$R^2$, all metrics         & $75.06\%(\pm0.22\%)$ & $74.97\%(\pm0.22\%)$ & $81.50\%(\pm0.23\%)$ & $81.48\%(\pm0.23\%)$ \\ 
$R^2$, entropy only        & $49.37\%(\pm0.32\%)$ & $49.02\%(\pm0.32\%)$ & $49.32\%(\pm0.31\%)$ & $49.12\%(\pm0.32\%)$ \\ 
\hline
\end{tabular}
}
\end{center}
\end{table}

From now on, we assign the term \emph{MetaSeg} to the introduced method. In \cite{Schubert2019} we extended this approach by taking resolution dependent uncertainty into account. As neural networks with their fixed filter sizes are not scale invariant, it makes a difference whether we infer the original input image or a resized one with the same network. Consequently, we introduced a pyramid-type of approach where a sequence of nested image crops with common center point are resized to a common size, than as a whole batch of input data inferred by the neural network, resized to their original size and than treated as an ensemble of predictions. Of this ensemble we can investigate mean and variance of dispersion measures and introduce further metrics, see~\cite{Schubert2019}. Due to this modification we gain roughly $3$ pp.\ for both meta tasks. A part of the effect is accounted to the introduction of resolution dependent uncertainty measures, while roughly an equal share stems from the deployment of neural networks for meta classification and regression.

\section{\label{sec:TDY}Time-Dynamic Meta Classification}

In online applications like automated driving, video streams of images are usually available. When inferring videos with single frame based convolutional neural networks, time dynamic uncertainties such as flickering segments can be observed. Therefore, as an extension of the previously introduced MetaSeg method, we present a time-dynamic approach for investigating uncertainties and assessing the prediction quality of neural networks over series of frames (meta regression) as well as for performing false positive detection (meta classification). In order to extend the single frame metrics to time series of metrics, we develop a light-weight tracking algorithm based on semantic segmentation, since by assumption the latter is already available. Segments in consecutive frames are matched according to their overlap in multiple frames. These measures are improved by shifting segments according to their expected location in the subsequent frame. By means of the identification of segments over time, we can extend each metric $M_i$ (defined as a scalar quantity in the previous section) to a time series. 
These time series are then presented to meta classifiers and regressors to perform both meta tasks.
The set of metrics used in \cite{Rottmann2018} is extended in \cite{Schubert2019}, this extension is deployed in the time-dynamic MetaSeg. A precise description of these metrics and of our tracking algorithm can be found in \cite{Maag2019}. All numerical tests in this section are performed using the updated set of metrics.

\begin{figure} 
    \subfigure[]{\includegraphics[width=0.24\textwidth]{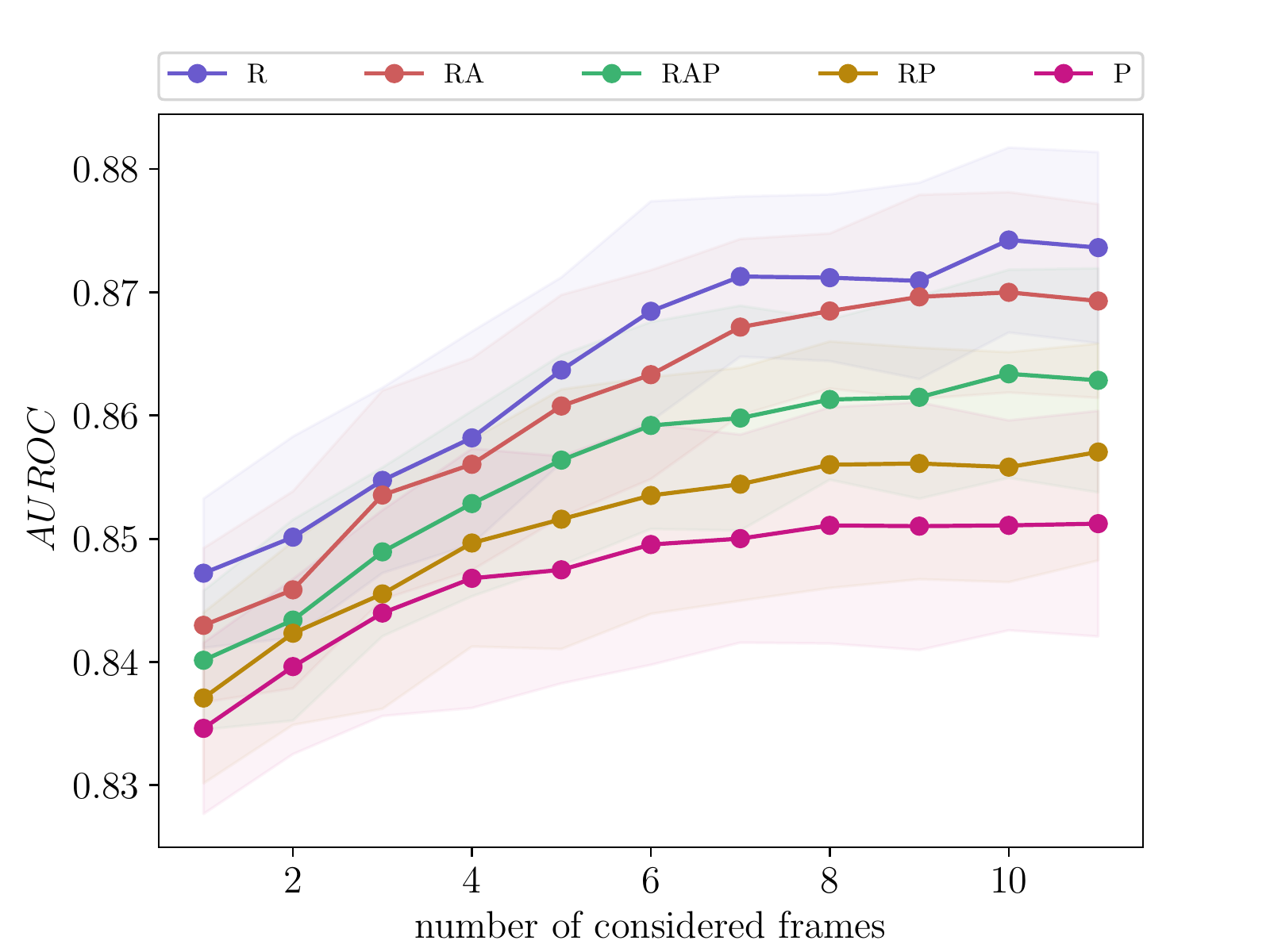}} 
    \subfigure[]{\includegraphics[width=0.24\textwidth]{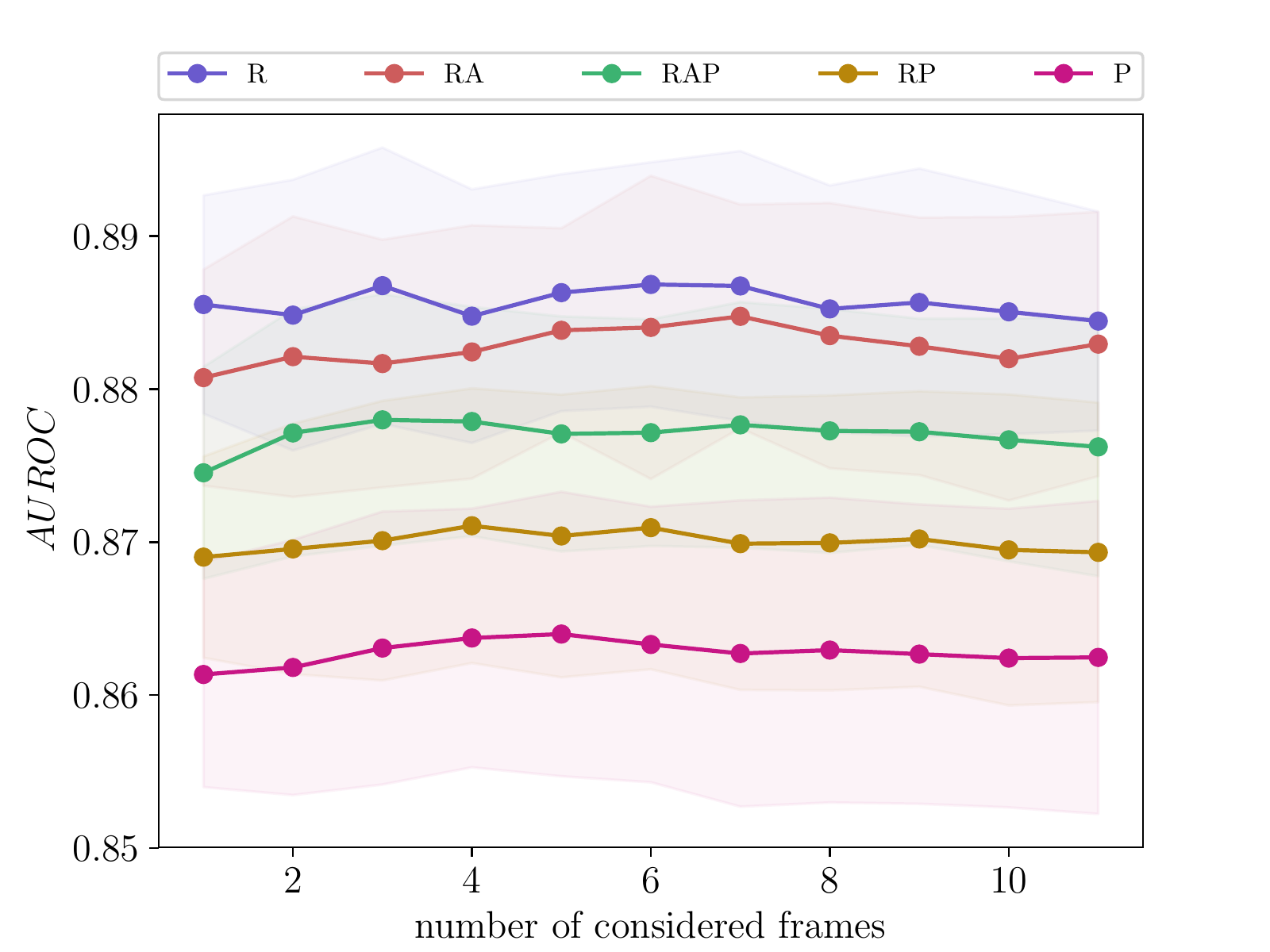}} 
\caption{A selection of results for meta classification AUROC as functions of the number of frames and for different compositions of training data. (a): meta classification via a neural network with $\ell_{2}$-penalization, (b): meta classification via gradient boosting.}
\label{fig:class_timeline}
\end{figure}

Let $\{ x_{1}, \ldots, x_{T} \}$ denote an image sequence with a length of $T$ and $x_{t}$ corresponds to the $t^\mathit{th}$ image. In what follows, we analyze the influence of the time series length on the models that perform meta classification and regression. In case when only using single frames (this corresponds to plain MetaSeg introduced in the previous section), we only present the segment-wise metrics $M^{t}$, where $M^{t}$ denotes the metrics of a single frame $t$, to the meta classifier/regressor. For the time dynamic approach, we extend the metrics to time series considering -- frame by frame -- up to $10$ previous frames and their metrics $M^{j}$, $j=t-10, \ldots, t-1$. In total, we obtain 11 different sets of metrics that are inputs for the meta classification and regression models. The presented results are averaged over 10 runs obtained by random sampling of the train/validation/test splitting. In \cref{fig:class_timeline} and \cref{tab:kitti}, the corresponding standard deviations are given by shades and in brackets, respectively. In addition to linear models used in the previous section, we also perform tests with gradient boosting and shallow neural networks with $\ell_{2}$-penalization for both meta tasks.

We perform tests with the KITTI dataset \cite{Geiger2013} containing street scene images from Karlsruhe, Germany. This dataset contains plenty of video sequences of which 29 contain ground truth. The tests we perform are based on these 29 sequences (yielding $\sim$12K images) containing 142 labeled (semanticly segmented) single frames in total. We use the same DeepLabv3+ networks like in the previous section (pre-trained on the Cityscapes dataset) to generate the output probabilities on the KITTI dataset. In our tests we mainly use the MobilenetV2 while the stronger Xception65 network serves as a reference network as to be explained subsequently. Since an evaluation of meta regression and classification requires a train/validation/test splitting, the small amount of 142 labeled images seems almost insufficient. Hence, we acquire alternative sources of useful information besides the (real) ground truth. First, we apply a variant of SMOTE for continuous target variables for \emph{data augmentation} (see \cite{Chawla2002,Torgo2013}) to augment the structured dataset of metrics. In addition, we utilize the Xception65 net with high predictive performance, its predicted segmentations we term \emph{pseudo ground truth}. We generate pseudo ground truth for all images where no ground truth is available. The train/val/test splitting of the data with ground truth available is 70\%/10\%/20\%. We use the alternative sources of information to create different compositions of training data, i.e., \textcolor{slateblue}{R} (real), \textcolor{indianred}{RA} (real and augmented), \textcolor{mediumseagreen}{RAP} (real, augmented and pseudo), \textcolor{darkgoldenrod}{RP} (real and pseudo) and \textcolor{mediumred-violet}{P} (pseudo). The shorthand ``real'' refers to ground truth obtained from a human annotator, ``augmented'' refers to data obtained from 
SMOTE and ``pseudo'' refers to pseudo ground truth obtained from the Xception65 net. These additions are only used during training. We utilize the Xception65 network exclusively for the generation of pseudo ground truth, all tests are performed using the MobilenetV2.

A selection of results for meta classification AUROC as functions of the number of frames, i.e., the maximum time series length, is given in \cref{fig:class_timeline}. The meta classification results for neural networks presented in subfigure (a) indeed show, that an increasing length of time series has a positive effect on meta classification. On the other hand, the results in subfigure (b) show that gradient boosting does not benefit as much from time series. In part this can be accounted to overfitting which we observe in our tests when using gradient boosting. 
Results for meta regression and meta classification are summarized in \cref{tab:kitti}. 
\begin{table}[t]
\begin{center}
\caption{Results for meta classification and regression for different compositions of training data and methods. The super script denotes the number of frames where the best performance and thus the given value is reached. The best results for each data composition are highlighted.}
\label{tab:kitti}
\scalebox{0.65}{
\begin{tabular}{||l||c|c||c|c||c|c||}
\cline{1-5}
\multicolumn{5}{||c||}{Meta Classification $\IoU=0,>0$} \\
\cline{1-5}
& \multicolumn{2}{c||}{Gradient Boosting} &\multicolumn{2}{c||}{Neural Network with $\ell_{2}$-penalization}   \\
\cline{1-5}
               & ACC    & AUROC  & ACC    & AUROC  \\ 
\cline{1-5}
\textbf{\textcolor{slateblue}{R}}         & $\mathbf{81.20}\boldsymbol{\%}(\pm1.02\%)^4$ & $\mathbf{88.68}\boldsymbol{\%}(\pm0.80\%)^6$ & $\mathbf{79.67}\boldsymbol{\%}(\pm0.93\%)^{10}$ & $\mathbf{87.42}\boldsymbol{\%}(\pm0.75\%)^{10}$ \rule{0mm}{3.5mm}    \\ 
\textbf{\textcolor{indianred}{RA}}        & $80.73\%(\pm1.03\%)^9$ & $88.47\%(\pm0.73\%)^7$ & $78.62\%(\pm0.61\%)^{11}$ & $87.00\%(\pm0.81\%)^{10}$     \\ 
\textbf{\textcolor{mediumseagreen}{RAP}}  & $79.64\%(\pm1.03\%)^7$ & $87.80\%(\pm0.82\%)^3$ & $77.08\%(\pm1.05\%)^9$ & $86.34\%(\pm0.84\%)^{10}$     \\ 
\textbf{\textcolor{darkgoldenrod}{RP}}    & $78.45\%(\pm0.88\%)^8$ & $87.11\%(\pm0.90\%)^4$ & $76.35\%(\pm0.67\%)^9$ & $85.70\%(\pm0.88\%)^{11}$     \\ 
\textbf{\textcolor{mediumred-violet}{P}}  & $77.56\%(\pm0.95\%)^5$ & $86.40\%(\pm0.93\%)^5$ & $75.68\%(\pm0.67\%)^{11}$ & $85.12\%(\pm0.92\%)^{11}$     \\ 
\cline{1-5}
\multicolumn{5}{||c||}{Meta Regression $\IoU$} \\ 
\cline{1-5}
& \multicolumn{2}{c||}{Gradient Boosting} &\multicolumn{2}{c||}{Neural Network with $\ell_{2}$-penalization}   \\
\cline{1-5}
       & $\sigma$ & $R^2$    & $\sigma$ & $R^2$  \rule{0mm}{3.2mm}   \\ 
\cline{1-5}
\textbf{\textcolor{slateblue}{R}}           & $0.114(\pm0.004)^5$ & $87.02\%(\pm1.00\%)^5$ & $\mathbf{0.113}(\pm0.005)^1$ & $\mathbf{87.16}\boldsymbol{\%}(\pm1.25\%)^1$ \rule{0mm}{3.5mm}    \\ 
\textbf{\textcolor{indianred}{RA}}          & $0.116(\pm0.004)^3$ & $86.39\%(\pm1.11\%)^3$ & $0.116(\pm0.005)^1$ & $86.46\%(\pm1.32\%)^1$    \\ 
\textbf{\textcolor{mediumseagreen}{RAP}}    & $\mathbf{0.112}(\pm0.003)^7$ & $\mathbf{87.51}\boldsymbol{\%}(\pm0.61\%)^7$ & $0.114(\pm0.005)^1$ & $86.97\%(\pm1.10\%)^1$    \\ 
\textbf{\textcolor{darkgoldenrod}{RP}}      & $\mathbf{0.112}(\pm0.002)^9$ & $87.45\%(\pm0.72\%)^9$ & $0.115(\pm0.003)^2$ & $86.69\%(\pm0.85\%)^2$  \\ 
\textbf{\textcolor{mediumred-violet}{P}}    & $0.114(\pm0.002)^{11}$ & $86.88\%(\pm0.67\%)^{11}$ & $0.117(\pm0.004)^3$ & $86.24\%(\pm0.99\%)^3$   \\ 
\cline{1-5}
\end{tabular} }
\end{center}
\end{table}
For gradient boosting as regression method we observe that the incorporation of pseudo ground truth slightly increases the performance. Noteworthily, we achieve almost the same performance when training gradient boosting either with pseudo ground truth exclusively or with real ground truth exclusively. This shows that meta regression can also be learned when there is no ground truth but a strong reference model available. We provide video sequences that visualize the $\IoU$ prediction and the segment tracking\footnote{ See \url{https://youtu.be/YcQ-i9cHjLk}}. For further results, especially those of the linear models, we refer to \cite{Maag2019}.
The results of the linear models are below those of gradient boosting in both meta tasks and are therefore not discussed in detail, here. 
In contrast to the single frame approach using only linear models, we increase the AUROC by $5.04$ pp.\ for meta classification and the $R^2$ by $5.63$ pp.\ for meta regression.

\section{\label{sec:DR}False Negative Detection by Decision Rules}

\begin{figure}
\begin{tikzpicture}
\node [align=center] at (.25\textwidth,0) {\includegraphics[width=.23\textwidth]{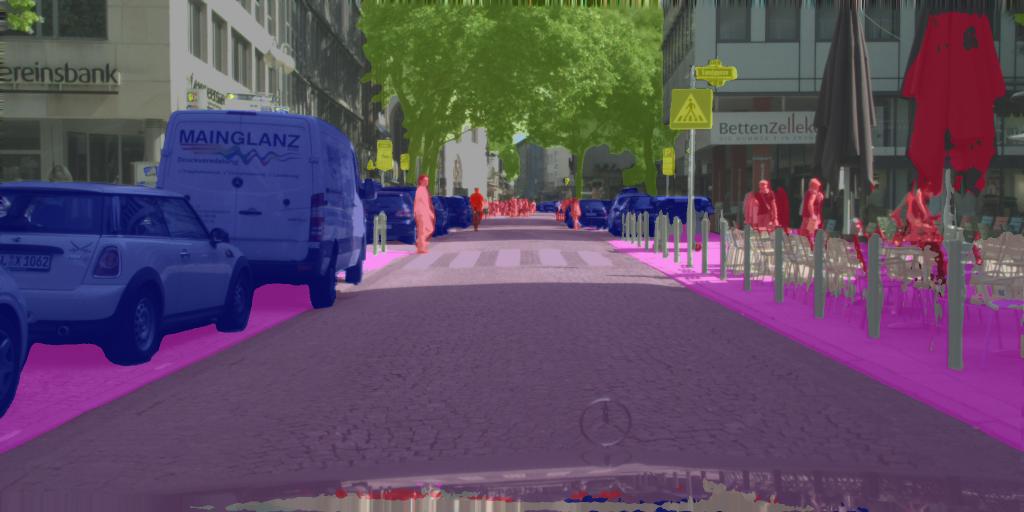}};
\node [align=center] at (.5\textwidth,0) {\includegraphics[width=.23\textwidth]{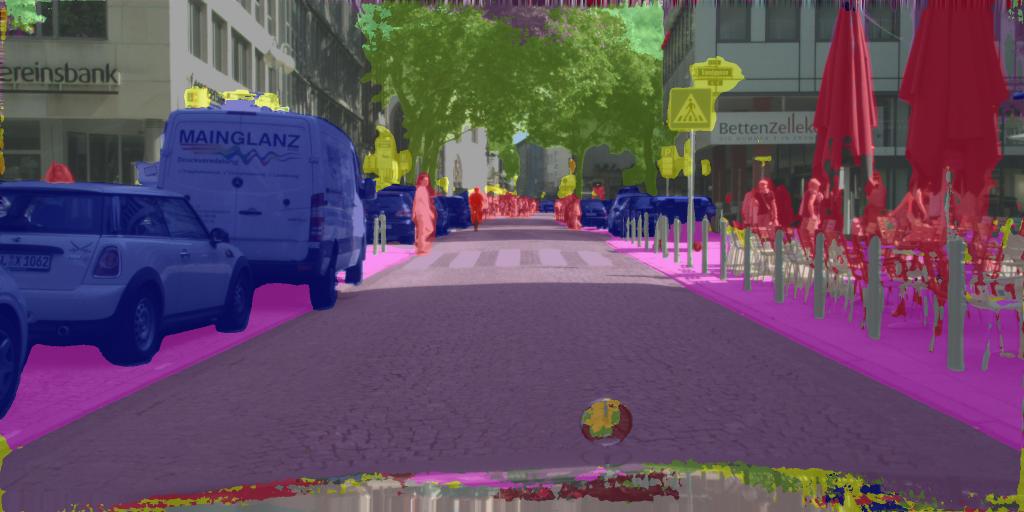}};
\node [align=center] at (.25\textwidth,1.4) {Bayes};
\node [align=center] at (.5\textwidth,1.4) {\phantom{y}Maximum Likelihood\phantom{y}};
\end{tikzpicture}
\caption{\label{fig:BayMLSeg}Illustration of two segmentation masks obtained with the Bayes decision rule (left) and the Maximum Likelihood decision rule (right).}
\end{figure}

\begin{figure}[t]
    \centering
    \vspace{.1in}
    \scalebox{.65}{\includegraphics[width=\linewidth, trim=0 .2in 0 .25in, clip]{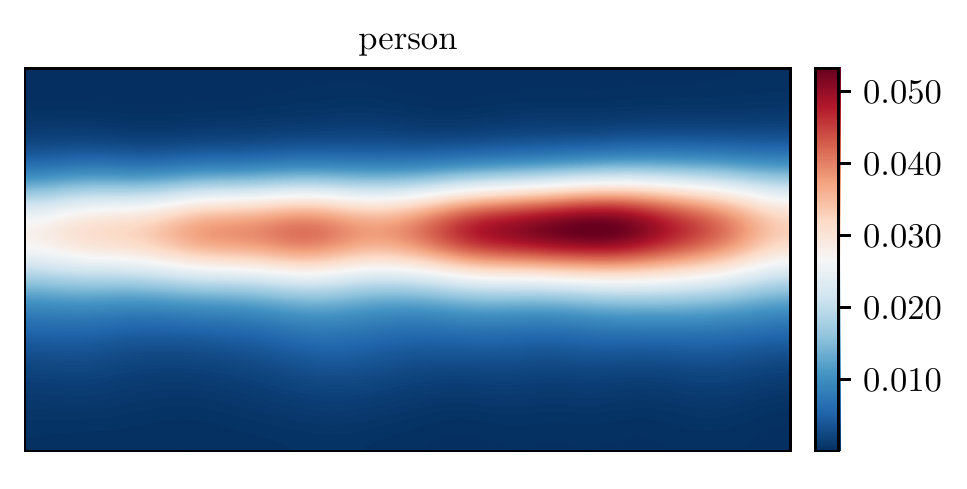}}
    \caption{Estimated pixel-wise prior probabilities of class \emph{human} in Cityscapes. For every other category, there is another heatmap with the property that the values at each pixel position over all heatmaps sum up to 1.}
    \label{fig:cs_prior_human}
\end{figure}

\begin{figure}[t]
\centering 
\scalebox{.75}{\includegraphics[width=\linewidth, trim=0 .2in 0 0]{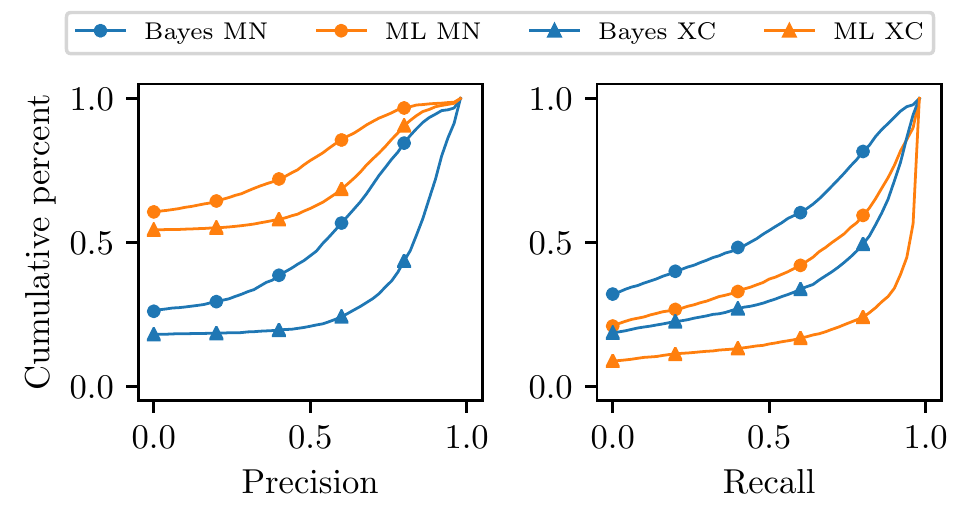}}
\caption{Empirical cumulative distribution functions in Cityscapes for segment-wise precision and recall of class \emph{human}.
} \label{fig:cs_cdf}
\end{figure}

In this section,
we draw 
attention to false-negative detection and the issue connected to the probabilistic output of segmentation networks when trained on unbalanced data, i.e., a dominant portion of pixels is assigned to only a few classes. As the softmax output of a segmentation network gives a pixel-wise class distribution over all $q$ predefined classes, the most commonly used decision rule, also known as maximum a-posteriori probability (MAP) principle, selects the class of highest probability. This is however merely one example of a cost-based decision rule and it is by far not the only possible selection principle. One could also penalize each confusion event by a specific quantity 
\begin{equation} \label{eq:cost-func}
    c_z\left(\hat{y},y\right) := 
    \begin{cases}
    0 &,\ \text{if}\quad \hat{y}=y \\
    \psi_z(\hat{y},y) &,\ \text{if}\quad \hat{y} \neq y
    \end{cases}\ ,\
    \psi_z(\hat{y},y) \in \mathbb{R}_{\geq0}
\end{equation}
that valuates the aversion of a decision maker towards the confusion of the predicted class $\hat y$ with the actual class $y$. The decision on the predicted class for pixel $z$ given image $x$ now minimizes the expected cost:
\begin{align}
    \hat{y}_z(x) 
    =&\ \argmin_{y^\prime \in \mathcal{C}} \mathbb{E}_z [ \ c_z(y^\prime,Y)\ |\ X=x\ ] \\
    =&\ \argmin_{y^\prime \in \mathcal{C}} \sum_{y\in\mathcal{C} \setminus \{ y^\prime \} } \psi_z(y^\prime,y) \, f_z(y|x) ~ .
\end{align} 
Seen from this angle, the standard MAP principle corresponds to cost functions that attribute equal cost to any confusion event, cf.~\eqref{eq:map}.
Although it seems reasonable, according to common human sense, to assume that $\psi_z(y^\prime,y)$ should be different depending on the type of confusion, another decision policy may reveal ethical problems when it comes down to providing explicit numbers~\cite{Chan2019Dilemma}. Therefore, the choice of cost functions to increase the sensitivity towards rare objects is subjected to constraints. A way out 
is offered by the
the mathematically appealing ``natural'' Maximum Likelihood (ML) decision rule which is known for its strength in finding instances of underrepresented classes in unbalanced datasets~\cite{Fahrmeir1996}. The latter rule assigns costs inverse proportional to the class frequencies, i.e.,
\begin{equation}
    \psi_z(y^\prime,y) = \frac{1}{\hat{p}_z(y)} = {|\mathcal{X}|} \left( \sum_{x \in \mathcal{X}} 1_{ \{y_z(x) = y\} } \right)^{-1} \forall ~ y^\prime \neq y
\end{equation}
with $\hat{p}_z(y)$ being the estimated a-priori probability (prior) from data $\mathcal{X}$ for class $y\in\mathcal{C}$ at location $z$. Considering 
a segmentation network as statistical model, the softmax output $f_z(y|x)$ can then be interpreted as a-posteriori probability of pixel $z$ in $x$ belonging to class $y$. Via the softmax adjustment with the priors 
\begin{align}
    \hat{y}_z(x)
    =& \argmin_{y^\prime \in \mathcal{C}} \sum_{y\in\mathcal{C} \setminus \{ y^\prime \} } \frac{1}{\hat{p}_z(y)} \, f_z(y|x) \\
    =& \argmax_{y \in \mathcal{C}} \frac{f_z(y|x)}{\hat{p}_z(y)} \stackrel{(\text{Bayes' Th.})}{=} \argmax_{y \in \mathcal{C}} f_z(x|y)
\end{align}
the class affiliation $y$ becomes an unknown parameter that needs to be estimated using the principle of maximum likelihood. The ML rule aims at finding the class $y$ for which the features $x$ are most typical, independent of any prior belief about the semantic classes such as the class frequency. We apply the ML rule in a position-specific manner in order to handle pixel-wise class imbalance, see~\cref{fig:BayMLSeg} and \cref{fig:cs_prior_human}.
Results for DeepLabv3+ (Xception65 and MobilenetV2) models on Cityscapes data are reported as empirical cumulative distribution functions (CDFs) of the category \emph{human} for segment-wise precision ($F^p$) and recall ($F^r$) in \cref{fig:cs_cdf}.


We observe an advantage of Bayes in terms of precision since $F_{\mathit{ML}}^p \prec F_{\mathit{B}}^p$ for both models, where $\prec$ stands for 1st order stochastic dominance \cite{pflug2007modeling} saying that typical precision values for Bayes are right shifted compared with ML. For any precision value $\mathit{v}$ 
, in particular for low precision values, the frequency 
with which an instance's precision is below $\mathit{v}$ 
is significantly less with Bayes than with ML.
In terms of recall, we observe the opposite behavior, i.e., ML is superior over Bayes in this metric. The steep ascent of the ML curves additionally indicates that most ground truth segments are predicted with high recall.
More relevantly, ML significantly reduces the number of non-detected segments, i.e., $F_{\mathit{B}}^r(0)>F_{\mathit{ML}}^r(0)$. Hence, the ML prediction can serve as uncertainty mask revealing image regions where an rare class object might be overlooked. For further reading, we refer to~\cite{Chan2019}.

\section{\label{sec:OL}Outlook}
The presented methods have clearly demonstrated their performance for false positive and false negative detection in semantic segmentation. Within this line of research, we plan to continue our work on constructing further time-dynamical metrics that quantify temporal uncertainty, as well as methods for false negative detection that do not overproduce false positives. Furthermore, transferring meta classification and regression to the task of object detection is a logical next step for future research.

Besides the mentioned algorithmic activities, several topics that can be considered as applications of false positive detection / false negative detection or segmentation quality assessment should be developed in the future. One application of meta regression is active learning for semantic segmentation. Here MetaSeg can be used as part of the query strategy.

As a future direction of reseach, work on out-of-distribution detection is a necessary next step to address the OOD failure mode.  We believe that the concept of meta classification and our method MetaSeg will play a significant role. Furthermore, we expect that the development of new uncertainty measures will play a crucial role \cite{gal2016dropout,oberdiek2018classification}. It is also an interesting question, in as much approaches of uncertainty quantification that guarantee OOD - detection can be transferred to machine learning with high dimensional input data \cite{meinke2019towards,hullermeier2019aleatoric}.

On the other hand, synthetic data should be integrated in this line of method development.  We expect that work on domain adaptation, active transfer learning methods as well as the generation of synthetic corner cases will play a vital role in the future. For all these applications, we believe that expressive uncertainty quantification and well-performing meta classification frameworks are key-components to establish new approaches or leverage existing ones.

Source codes for our frameworks are available on GitHub, see \url{https://github.com/mrottmann/MetaSeg}.

\bibliographystyle{IEEEtran}
\bibliography{biblio}
\end{document}